\documentclass[runningheads]{llncs}
\usepackage[T1]{fontenc}

\usepackage{graphicx}
\usepackage{amsmath}
\usepackage{booktabs}
\usepackage{caption}
\usepackage{subcaption}
\usepackage{multirow}
\usepackage{multicol}
\usepackage{hyperref}
\usepackage{xcolor}
\usepackage{comment}
\usepackage{fancyhdr}

\definecolor{faded}{HTML}{9e9e9e}

\renewcommand{\rightmark}{Revisiting Document Image Classification Using LLMs}

\begin{document}
\title{Zero-Shot Prompting and Few-Shot Fine-Tuning: Revisiting Document Image Classification Using Large Language Models}
\titlerunning{Revisiting Document Image Classification Using LLMs}

\author{%
Anna Scius-Bertrand\thanks{These authors contributed equally to this work.}\inst{1, 2}%\orcidID{0009-0006-7414-2214}
\and
Michael Jungo\textsuperscript{$\star$}\inst{1, 2}%\orcidID{0009-0001-1790-1687}
\and
Lars Vögtlin\inst{2}%\orcidID{0000-0002-2543-9074}
\and
Jean-Marc Spat \inst{1}
\and
Andreas Fischer\inst{1, 2}%\orcidID{0000-0003-0069-3436}
}

\authorrunning{A. Scius-Bertrand and M. Jungo et al.}

\institute{University of Applied Sciences and Arts Western Switzerland \\
\email{\{anna.scius-bertrand,michael.jungo,jean-marc.spat,andreas.fischer\}@hefr.ch}  \and
University of Fribourg, Switzerland \\
\email{\{anna.scius-bertrand,michael.jungo,lars.vogtlin,andreas.fischer\}@unifr.ch}}

\maketitle              % typeset the header of the contribution

\begin{abstract}
Classifying scanned documents is a challenging problem that involves image, layout, and text analysis for document understanding. Nevertheless, for certain benchmark datasets, notably RVL-CDIP, the state of the art is closing in to near-perfect performance when considering hundreds of thousands of training samples. With the advent of large language models (LLMs), which are excellent few-shot learners, the question arises to what extent the document classification problem can be addressed with only a few training samples, or even none at all. In this paper, we investigate this question in the context of zero-shot prompting and few-shot model fine-tuning, with the aim of reducing the need for human-annotated training samples as much as possible.


\end{abstract}

\thispagestyle{fancy}
\renewcommand{\headrulewidth}{0pt}
\lfoot{\scriptsize{\textit{Proceedings of the $27^{th}$ International Conference on Pattern Recognition 2024 (ICPR 2024)}}}

\section{Introduction}

The RVL-CDIP dataset, introduced by Harley et al.~\cite{harley2015icdar}, is a subset of 400\,000 labeled document images derived from the IIT-CDIP collection, originating from a litigation against the tobacco industry~\cite{lewis2006building}, which has significantly boosted the exploration of deep learning methods for document image classification in the last decade. Notable advancements include the application of convolutional neural networks~\cite{tensmeyer2017analysis}, the integration of text, image, and layout embeddings as demonstrated by LayoutLM~\cite{xu2021layoutlmv2}, OCR-free document understanding through Transformers (Donut)~\cite{kim2022donut}, and cross-modal strategies that fuse image and textual analysis techniques~\cite{bakkali2020visual}. The cross-modal strategies stand out by achieving an impressive classification accuracy of 97.05\% across 16 distinct document types, including letters, forms, and emails.

Nevertheless, neural network models with a rising number of trainable parameters require a large training set of labeled documents to perform satisfactorily. For example, the RVL-CDIP benchmark needs a training set of 320\,000 labeled documents to distinguish between 16 document classes. If the document categories change, or for a new dataset, the training set must be relabeled accordingly, which leads to a costly and time-consuming human effort.

The recent advent of large language models (LLMs) has impressively shown that large networks with billions, or even more than a trillion, parameters, only need very few training samples, if any, to solve challenging tasks in natural language processing, including closed-book question answering, translation, and reading comprehension~\cite{brown2020language}. LLMs typically rely on unsupervised pre-training of decoder-only transformer architectures on a large body of texts from the internet, such as the Common Crawl dataset~\cite{raffel2020exploring}, followed by reinforcement learning from human feedback, to achieve astounding generalization capabilities.

With respect to the task of document classification, the question arises to what extent LLMs may be capable of solving the task without the need of hundreds of thousands of learning samples, relying on their text understanding capabilities of the document texts, which are extracted by means of optical character recognition (OCR).

In the present paper, we explore this question in a comprehensive benchmark evaluation that takes into account several state-of-the art LLMs for text analysis (Mistral~\cite{jiang2023mistral}, GPT-3~\cite{brown2020language}, GPT-4~\cite{achiam2023gpt}), defines different training scenarios, and puts the LLM results into a broader context by including also a selection of smaller language models (RoBERTa~\cite{roberta}), text embedding models (Jina~\cite{gunther2023jina}), OCR-free image-based models (Donut~\cite{kim2022donut}), and multi-modal LLMs (GPT-4-Vision~\cite{achiam2023gpt}) in the comparison. The aim of the benchmark evaluation is to investigate the document classification performance for an increasing number of learning samples, starting with zero-shot prompting, where only a textual description of the task is provided to the model, and ending with few-shot model fine-tuning using 100 samples per class. Note that fine-tuning of LLMs is a challenging task on standard hardware. We rely on Low-Rank Adaptation (LoRA)~\cite{lora} to fine-tune one of the smaller open source LLMs, Mistral-7B~\cite{mistral}, for the purpose of our benchmark.

No new method is proposed in this paper. Instead, our contributions are:
\begin{itemize}
\item A comprehensive benchmark\footnote{We will make the benchmark data and models publicly available with the publication of the paper.} for evaluating document classification in a few-shot training scenario.
\item Comparing LLM prompting vs LLM fine-tuning for document classification.
\item Comparing generative vs embedding-based document classification.
\item Comparing text-based vs image-based document classification.
\end{itemize}
With this benchmark evaluation and comparisons of current interest, we aim to inspire and support further research on few-shot document classification.

The paper is structured as follows: An introduction of the dataset is given in \autoref{sec:db}, a description of the methods in \autoref{sec:method}, and in \autoref{sec:expe} the experimental results are presented. Lastly, we draw some conclusions and discuss future work.

\section{Data}
\label{sec:db}

\begin{figure}
    \centering
    \includegraphics[width=\linewidth]{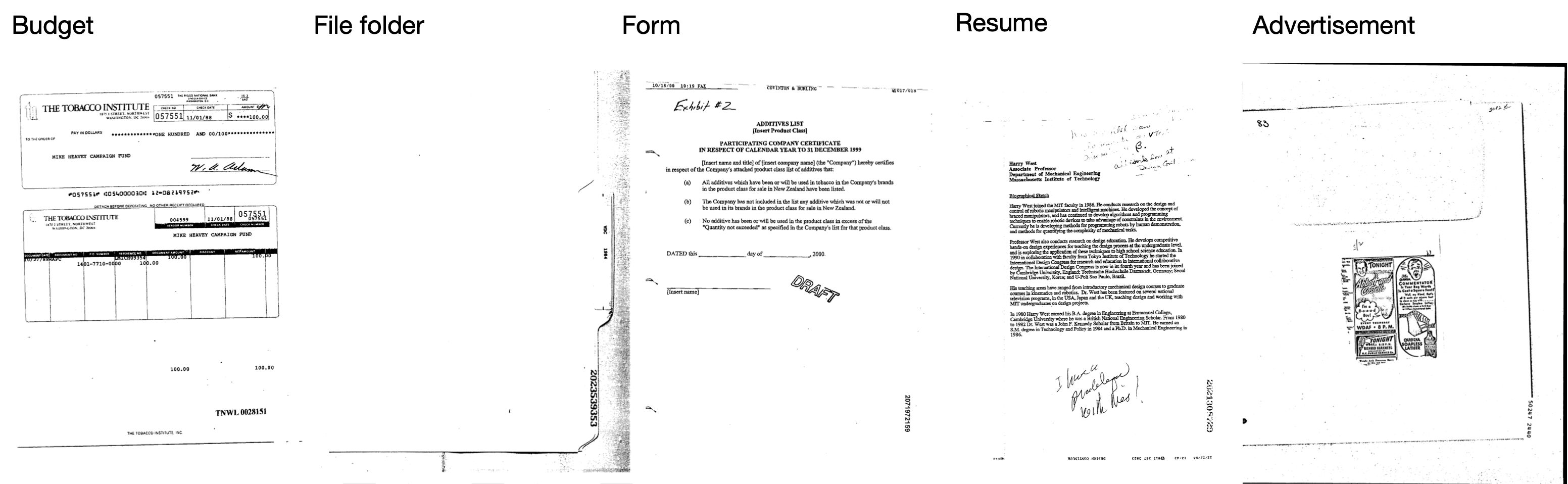}
    \caption{Example document images from the RVL-CDIP dataset.}
    \label{fig:samples}
\end{figure}

The RVL-CDIP dataset~\cite{harley2015icdar} contains 25\,000 scanned grayscale images per document class for 16 classes: letter, form, email, handwritten, advertisement, scientific report, scientific publication, specification, file folder, news article, budget, invoice, presentation, questionnaire, resume, and memo. Several examples from different categories are shown in \autoref{fig:samples}.

Many document classification and understanding methods require the text contained in the images, which can be obtained with any OCR engine. No matter how good the OCR engines have become, there are still difficulties to obtain a perfect textual representation of the documents. Most notably, parts of some images are illegible due to degradation or any other quality issues. This makes purely text-based methods more difficult, as they also need to deal with the noise created by the OCR.

Furthermore, several categories contain images with very little textual information, making it potentially more difficult to distinguish between them as file folders or advertisements shown in~\autoref{fig:samples}. There is also a mix between handwriting and printed text in the documents, which could cause issues for the OCR. In any case, the visual information that is attached to the different types of writing is lost in the process. Therefore, any OCR-based method relies much more heavily on the meaning of the content, rather than its structure, although the grammatical structure remains a significant part of the textual representation.

\section{Methods}
\label{sec:method}

In the following, we describe the different methods used in the benchmark: Text-based classification using generative LLMs, text embedding classification, image-based document classification, and multi-modal LLMs.

\subsection{Text-Based Classification}

\subsubsection{OCR}

For OCR, we use Amazon's Textract, which performed reasonably well on a few example images from the RVL-CDIP dataset, especially with respect to low-resolution, skewed documents, and handwritten elements. The resulting machine-readable text follows a natural reading order and the text lines are delimited by line-break characters.

\subsubsection{LLMs}

We focus on some of the best-performing LLMs from the current state of the art, including the GPT models from OpenAI~\cite{achiam2023gpt,brown2020language,radford2018improving,radford2019language} and the models from Mistral AI~\cite{jiang2023mistral}. The transformer-based models have been pretrained on texts from the internet and fine-tuned to follow instructions, as well as using reinforcement learning from human feedback. 

\begin{table}[ht]
    \setlength{\tabcolsep}{10pt}
    \def\arraystretch{1.3}
    \centering
    \caption{LLM versions.}\label{tab:llms}
    \vspace{0.5em}
    \begin{tabular}{l l}
          \toprule
          Model  & Version \\
          \midrule
          Mistral-7B & Open source~\cite{jiang2023mistral} \\
          Mixtral-8x7B & Open source~\cite{jiang2024mixtral} \\
          Mistral-Medium & mistral-medium-2312 \\
          Mistral-Large & mistral-large-2402 \\
          GPT-3.5 & gpt-3.5-turbo-0125 \\
          GPT-4 & gpt-4-turbo-2024-04-09 \\
          \bottomrule
    \end{tabular}
\end{table}

\autoref{tab:llms} lists the specific versions used. Mistral-7B has $7 \times 10^{9}$ parameters~\cite{jiang2023mistral}, Mixtral-8x7B has $47 \times 10^{9}$~\cite{jiang2024mixtral}, and GPT-3 has $175 \times 10^{9}$~\cite{brown2020language}. To the best of our knowledge, no official information is available at the moment for the larger models.

\subsubsection{LLM Prompts}

The LLMs are used in chat completion mode with a system prompt that specifies the classification task in natural language. The prompt was first formulated by a human. Afterwards, we used GPT-4 in chat completion mode to refine the original prompt and used it again to further tweak the refined prompt. The three resulting prompts are shown in \autoref{fig:prompts}.

\begin{figure}
  \centering
  \begin{subfigure}[t]{0.32\textwidth}
    \captionsetup{justification=centering}
    \caption*{\textbf{Prompt 1 (P1)}\\Written by a human}
    \vspace{0.5em}
    \scriptsize{\texttt{%
      You are a specialist for classifying documents based on their OCR text. I will show you the OCR text of a document. There are only 16 classes:\\
      {[}"letter", "form", [\dots], "resume", "memo"]\\
      Choose one of these 16 classes, only a single class, and only from the list. Do not comment your result, I only want to know the name of the class.
    }}
  \end{subfigure}\hfill%
  \begin{subfigure}[t]{0.32\textwidth}
    \captionsetup{justification=centering}
    \caption*{\textbf{Prompt 2 (P2)}\\P1 improved by GPT4}
    \vspace{0.5em}
    \scriptsize{\texttt{%
      Your task is to classify a document's content based solely on its OCR text. Select one appropriate category for the document from the following 16 options:\\\\
      1. letter\\
      2. form\\
      {[}\dots]\\
      15. resume\\
      16. memo\\\\
      Provide only the chosen category name from the list as your response, without any additional commentary or explanation.
    }}
  \end{subfigure}\hfill%
  \begin{subfigure}[t]{0.32\textwidth}
    \captionsetup{justification=centering}
    \caption*{\textbf{Prompt 3 (P3)}\\P2 tweaked with GPT4}
    \vspace{0.5em}
    \scriptsize{\texttt{%
      Classify the content of a document using its OCR text. Choose exactly one category from the list below that best fits the document:\\\\
      - letter\\
      - form\\
      {[}\dots]\\
      - resume\\
      - memo\\\\
      Respond with the name of the selected category only. Do not provide additional comments or explanations.
      }}
  \end{subfigure}
  \caption{Three system prompts considered for document classification. Note that the list of 16 categories has been shortened in the figure, indicated by [\dots], to save space.}\label{fig:prompts}
\end{figure}

\paragraph{Zero-shot prompting:} Only the system prompt is used, relying on the semantics of the category names for classification.

\paragraph{One-shot prompting:} We provide context to the LLM by including 16 additional pairs of prompts ($user_i$, $assistant_i$) after the system prompt, one for each category, where $user_i$ is an OCR text from the training set and $assistant_i$ is its category name.

\subsubsection{Fine-Tuning LLMs}

LLMs have learned a deep understanding of text through an extensive pretraining. Even though their primary appeal is to use them generatively, which makes them extremely flexible, it would stand to reason that their capabilities can also serve as a base for a classification model. In particular, understanding documents in a low-resource context could benefit from their vast knowledge of all kinds of texts, including a large variety of documents.
Hence, we explore the task of document classification by adding a classifier on top of an LLM. In order to fine-tune the LLM, we further employ Low-Rank Adaptation (LoRA)~\cite{lora}, to adapt it to the specific documents at hand.

As a comparison, to evaluate the effectiveness of using an LLM as the base model, we also fine-tune the largest available pretrained RoBERTa~\cite{roberta}, which only has 355M parameters. Because of its comparatively small size, the model is fine-tuned fully instead of having to resort to LoRA adapters.

Finally, to see whether adding a classifier head is really necessary, we also fine-tune the same model in a generative manner, where at the end of each document, a classification instruction is added. With the generative approach, the model maintains the ability to be adapted to any kind of output, as opposed to the classifier, where the output is limited to the classes it has been trained for. There are a few disadvantages that come with this flexibility, most notably that the output is no longer guaranteed to be exactly the class that was asked for, and an increase in compute, as the model now needs to predict multiple tokens. Even when the class name consists of a single token, the generative model needs to predict at least an end-of-sequence token in addition to the class name, which requires multiple forward passes.

\subsection{Embedding-Based Methods}

In addition to generative approaches for classifying OCR texts, we also consider a standard classification setup based on feature vector representations of the OCR texts in an embedding space with a $k$-nearest neighbor classification (KNN).
We also conducted preliminary experiments with multi-layer perceptions (MLP), the results of which were close to the KNN but never outperformed it. 

We focus on some of the best-performing text embedding models from the current state of the art, specifically a selection of models proposed by Jina AI~\cite{gunther2023jina}, Mistral AI, and OpenAI~\cite{neelakantan2022text}. The versions used are listed in \autoref{tab:embs}.

\begin{table}[ht]
    \setlength{\tabcolsep}{10pt}
    \def\arraystretch{1.3}
    \centering
    \caption{Embedding model versions.}\label{tab:embs}
    \vspace{0.5em}
    \begin{tabular}{l l r}
          \toprule
          Model  & Version & Embedding size \\
          \midrule
          Jina-v2 & jina-embeddings-v2-base-en & 768 \\
          Mistral-embed & mistral-embed & 1024 \\
          OpenAI-small & text-embedding-3-small & 1536 \\
          OpenAI-large & text-embedding-3-large & 3072 \\
          \bottomrule
    \end{tabular}
\end{table}

\subsection{Image-Based Methods}

\subsubsection{Donut}

Donut is a Transformer-based model for Visual Document Understanding (VDU) that operates entirely on images, without having to rely on the OCR. Besides a large number of synthetic documents, the pretraining also included the full IIT-CDIP dataset, which makes it particularly well-suited for classifying documents from the RVL-CDIP subset.

In \cite{kim2022donut}, they already fined-tuned Donut on various downstream tasks, including the classification of RVL-CDIP, which showed excellent results. However, this was conducted on the complete RVl-CDIP dataset with 320K images in the training set, whereas we are investigating whether it also performs well when the training data is severely limited.

\subsection{Multi-Modal Methods}

\subsubsection{GPT-4-Vision}
Recently, GPT-4-Vision was introduced as a multi-modal extension of GPT-4, which accepts a combination of text and/or image inputs from the user~\cite{achiam2023gpt} (with the same model version as indicated in \autoref{tab:llms}).

\paragraph{Image-based zero-shot prompting.} The first sentence of the system prompts listed in \autoref{fig:prompts} is changed to ``Your task is to classify a document.'' without mentioning the OCR text. Only the document image is provided as input.

\paragraph{Bimodal zero-shot prompting.} The first sentence of the system prompts is changed to ``Your task is to classify a document based on its OCR text and scanned image, which are both provided by the user.'' Afterwards, both the OCR text and the document image are provided as input.

\section{Experimental Evaluation}
\label{sec:expe}

In this section, we first describe the experimental setup of the benchmark, followed by results obtained for prompting, embedding, and fine-tuning, respectively. At the end, all results are summarized and put into context with other results from the current state of the art.

\subsection{Setup}
\label{sec:expe_setup}

\subsubsection{Training.} For zero-shot and one-shot prompting, we consider zero samples and 16 samples (one per class), respectively. For few-shot fine-tuning, we investigate an increasing number of 160 samples (ten per class), 800 samples (50 per class), and 1\,600 samples (100 per class), which are randomly chosen from the original RVL-CDIP training set. Note that the smaller training sets are included in the larger ones.

\subsubsection{Validation.} For optimizing hyperparameters, we consider a validation set of an additional 160 samples (ten per class), which are randomly chosen from the original RVL-CDIP validation set.

\subsubsection{Testing.}

We define a test scenario \textbf{RVL-CDIP-160x5}, which consists of five random selections (without overlap) of 160 samples (ten per class) from the original test set, for which we obtain a high-quality OCR. We report the mean accuracy and standard deviation over the five test sets. As demonstrated in the experiments (see~\autoref{sec:expe_finetuning}), this selection results in a representative subset, which reduces the cost of testing significantly when using the APIs of Mistral AI and OpenAI for their LLMs, as they are not freely available.

Additionally, the original \textbf{RVL-CDIP-40K} test set, which contains 40\,000 samples, is also included in the evaluation, whenever possible. Having this large test set available, increases the confidence in the perceived evaluation and serves as a reference to existing results from the literature.

\subsubsection{OCR.}

For the training sets, validation set, and the RVL-CDIP-160x5 test sets, we extract the text from the images by using Amazon's Textract.
As running the OCR on the full RVL-CDIP-40K test set goes beyond the scope of this research, we fall back to the original OCR texts from the IIT-CDIP collection~\cite{lewis2006building}. The original OCR was performed with a 90s-era OCR engine, which unquestionable produced a lower quality output. Nevertheless, we include it to judge how representative our randomly selected subsets are with respect to an established test set of considerable size.

\subsubsection{Models setup.}

For zero-shot and one-shot prompting, we rely on the APIs of Mistral AI and OpenAI respectively, specifying a temparature of $0$ to encourage precise, non-creative answers. The LLM versions used are indicated in~\autoref{tab:llms}.

For KNN-based classification of OCR embeddings, we use the APIs of Jina AI, Mistral AI, and OpenAI to extract the embedding vectors. The embedding models considered are indicated in~\autoref{tab:embs}. The parameter $k \in \{1, 3, 5, 7, 9\}$ and the metric (Euclidean or cosine) are optimized with respect to the classification accuracy on the validation set.

Fine-tuning Donut is performed by using the official implementation\footnote{\url{https://github.com/clovaai/donut}} and with the suggested configuration of the CORD dataset~\cite{park2019cord}. In particular, the image size is fixed to $1280 \times 960$ pixels, which provides a good compromise between readability for the human eye and computational effort for the GPU.

Mistral-7B\cite{mistral} is used as the base model for all experiments for the LLM fine-tuning. As proposed in QLoRA~\cite{qlora}, the base model is quantized into 4-bit weights and trainable LoRA adapters are added to all linear layers. We chose rank $r = 8$ with $alpha = 16$ based on preliminary experiments on the validation set.
On the other hand, for RoBERTa, all parameters are fine-tuned, since the model is small enough to be trained fully in a reasonable time on the available hardware.

The generative fine-tuning adds a classification directive at the end of each document, specifically ``\texttt{\#\#\# Classification:}'' followed by the class name to be predicted. Solely the tokens after the classification directive contribute to the loss and therefore the weight updates. This is not an instruction tuning, meaning that the model does not receive the instructions that were used for the one-shot prompting.

RoBERTa and Mistral-7B are both implemented and trained with HuggingFace's \texttt{transformers}~\cite{hf-transformers} library and \texttt{bitsandbytes}\footnote{\href{https://github.com/TimDettmers/bitsandbytes}{https://github.com/TimDettmers/bitsandbytes}} to support QLoRA.

\subsection{Prompting Results}
\label{sec:expe_prompting}

In a preliminary experiment on the validation set, we optimized the LLM system prompt (see ~\autoref{fig:prompts}): Written by a human (P1), enhanced by GPT-4 (P2), and enhanced twice by GPT-4 (P3). For a few-shot learning task with GPT-3.5 using 32 training samples (two per class) and evaluating on 48 validation samples (three per class), the system prompt P2 achieved the best accuracy (60.4\%), outperforming P1 (58.3\%) and P3 (56.2\%). That is, an enhancement of the human prompt by GPT-4 was beneficial and P2 was selected for all subsequent experiments.

\begin{table}[ht]
    \setlength{\tabcolsep}{10pt}
    \def\arraystretch{1.3}
    \centering
    \caption{\textbf{Zero-shot and one-shot prompting.} Document classification results on the RVL-CDIP-160x5 test sets, indicating the number of training samples~(\#Train), as well as the mean and standard deviation of the classification accuracy~(\%) and invalid answers~(\%) across the five subsets. The best results per training scenario are marked in \textbf{bold}.}\label{tab:prompting}
    \vspace{0.5em}
    \begin{tabular}{r l l | r r}
\toprule
\#Train & Input & Model & Accuracy & Invalid\\
\midrule
0 & OCR & Mistral-7B & 45.4 $\pm$ 2.8 & 17.0 $\pm$ 2.9 \\
0 & OCR & Mixtral-8x7B & 25.0 $\pm$ 2.9 & 56.2 $\pm$ 1.7 \\
0 & OCR & Mistral-Medium & 54.6 $\pm$ 4.1 & 6.4 $\pm$ 3.1 \\
0 & OCR & Mistral-Large & 54.4 $\pm$ 4.1 & 14.6 $\pm$ 1.6 \\
0 & OCR & GPT-3.5 & 36.9 $\pm$ 1.1 & 32.1 $\pm$ 1.5 \\
0 & OCR & GPT-4 & 61.8 $\pm$ 2.0 & 2.1 $\pm$ 1.2 \\
0 & Image & GPT-4-Vision & \textbf{69.9} $\pm$ \textbf{2.0} & 0.5 $\pm$ 0.5 \\
0 & OCR+Image & GPT-4-Vision & 69.4 $\pm$ 1.7 & 0.8 $\pm$ 0.7 \\
\midrule
16 & OCR & Mistral-7B & 47.1 $\pm$ 3.7 & 22.9 $\pm$ 2.8 \\
16 & OCR & Mixtral-8x7B & 48.2 $\pm$ 5.9 & 13.4 $\pm$ 3.3 \\
16 & OCR & GPT-3.5 & \textbf{58.8} $\pm$ \textbf{2.1} & 4.6 $\pm$ 1.8 \\
\bottomrule
    \end{tabular}
\end{table}

\autoref{tab:prompting} shows the prompting results on the test sets of RVL-CDIP-160x5. For zero-shot prompting, the mean classification accuracy ranges from 25.0\% (Mixtral-8x7B) to 69.9\% (GPT-4-Vision), highlighting a large variability among the models.

GPT-4-Vision significantly outperforms GPT-4, highlighting the importance of the document image for classification. Bimodal prompting with both OCR text and image did not further improve the results when compared with image-only prompting. Note, however, that GPT-4-Vision is capable of performing OCR, at least implicitly, when the input consists only of the document image.

One of the main limitations of the smaller models (Mistral-7B, Mixtral-8x7B, GPT-3.5) are invalid answers. Instead of only responding with a category name, as requested in the prompt, the models tend to produce longer responses, e.g. ``The text provided appears to be a notice for a membership investment in the Florida Retail Political Action Committee''. We do not post-process the responses, thus any deviation from valid category names is considered invalid.

The one-shot prompting results show that the mean accuracy can be improved for the smaller models (Mistral-7B, Mixtral-8x7B, GPT-3.5) by providing one example per class, in particular Mixtral-8x7B is improved from 25.0\% to 48.2\% while reducing the number of invalid answers from 56.2\% to 13.4\%.

We did not test one-shot prompting for the larger models due to the large number of tokens that are added to the prompt (16 OCR texts or images, respectively), which significantly increases the costs of using the commercial APIs.

\subsection{Embedding Results}
\label{sec:expe_embedding}

The results for KNN-based classification of OCR text embeddings are shown in \autoref{tab:knn-embeddings}. The mean accuracy achieved by Mistral-embed, OpenAI-small, and OpenAI-large are fairly similar and outperform most of the LLM prompting results when $800$ or more training samples are considered, demonstrating that embeddings are a promising strategy for a few-shot learning scenario.

\begin{table}[ht]
    \setlength{\tabcolsep}{10pt}
    \def\arraystretch{1.3}
    \centering
    \caption{\textbf{KNN classification of OCR text embeddings}. Document classification results on the RVL-CDIP-160x5 test sets, indicating the number of training samples~(\#Train), as well as the mean and standard deviation of the classification accuracy~(\%) across the five subsets. The best results per training scenario are marked in \textbf{bold}.}\label{tab:knn-embeddings}
    \vspace{0.5em}
    \begin{tabular}{r l l r}
\toprule
\#Train & Input & Embedding & Accuracy\\
\midrule
160 & OCR & Jina-v2 & 41.9 $\pm$ 1.6 \\
160 & OCR & Mistral-embed & \textbf{56.4} $\pm$ \textbf{3.0} \\
160 & OCR & OpenAI-small & 53.9 $\pm$ 2.9 \\
160 & OCR & OpenAI-large & 54.4 $\pm$ 1.2 \\
\midrule
800 & OCR & Jina-v2 & 52.9 $\pm$ 4.0 \\
800 & OCR & Mistral-embed & 63.5 $\pm$ 2.0 \\
800 & OCR & OpenAI-small & 62.4 $\pm$ 3.2 \\
800 & OCR & OpenAI-large & \textbf{64.8} $\pm$ \textbf{3.1}\\
\midrule
1\,600 & OCR & Jina-v2 & 57.1 $\pm$ 4.5 \\
1\,600 & OCR & Mistral-embed & 66.8 $\pm$ 2.7 \\
1\,600 & OCR & OpenAI-small & 65.8 $\pm$ 3.4 \\
1\,600 & OCR & OpenAI-large & \textbf{67.8} $\pm$ \textbf{3.8} \\
\bottomrule
    \end{tabular}
\end{table}

There is one exception: Zero-shot prompting using GPT-4-Vision ($69.9\%$ mean accuracy) outperforms all embedding models tested, even when using $1\,600$ training samples. 

\begin{figure}
\centering
\includegraphics[width=0.9\textwidth]{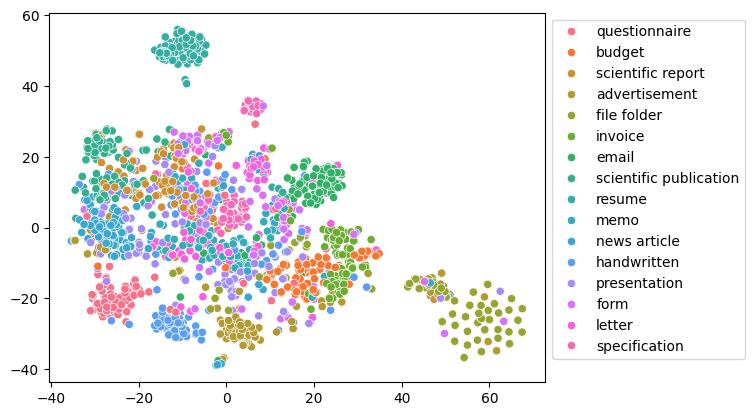}
\caption{Embedding of the $1\,600$ training samples using the OpenAI-large model, visualized with t-SNE.\label{fig:embs}}
\end{figure}

A visualization of the OpenAI-large embeddings using t-SNE dimensionality reduction is depicted in \autoref{fig:embs}, illustrating the capability of the embedding model to form class-wise clusters based on the OCR text.

\subsection{Fine-Tuning Results}
\label{sec:expe_finetuning}

\autoref{tab:finetuning} reports the results for fine-tuning RoBERTa, Mistral-7B, and Donut in the different few-shot training scenarios. All fine-tuned models outperform LLM prompting and embedding strategies from the previous sections when $800$ training samples or more are used. The results also indicate that almost no invalid responses are generated by the fine-tuned models.

\begin{table}[ht]
  \setlength{\tabcolsep}{5pt}
  \def\arraystretch{1.2}
  \centering
  \caption{\textbf{Few-shot model fine-tuning.} Document classification results on the RVL-CDIP-160x5 and RVL-CDIP-40K test sets, indicating the number of training samples~(\#Train) and the classification accuracy~(\%) across the five subsets. The best results per training scenario are marked in \textbf{bold}.}\label{tab:finetuning}
  \vspace{0.5em}
  \begin{tabular}{r l l | c c | c c}
    \toprule
    & & & \multicolumn{2}{c}{RVL-CDIP-160x5} & \multicolumn{2}{|c}{RVL-CDIP-40K} \\
    \#Train & Input & Model & Accuracy & Invalid & Accuracy & Invalid \\
    \midrule
    160 & OCR & RoBERTa & 59.8 $\pm$ 3.8 & 0.0 $\pm$ 0.0 & 50.2 & 0.0 \\
    160 & OCR & Mistral-7B-Class & 51.1 $\pm$ 4.0 & 0.0 $\pm$ 0.0 & 23.4 & 0.0 \\
    160 & OCR & Mistral-7B-Gen & \textbf{72.5} $\pm$ \textbf{3.9} & 0.4 $\pm$ 0.3 & 66.7 & 0.2 \\
    160 & Image & Donut & 42.8 $\pm$ 3.0 & 0.8 $\pm$ 0.7 & 44.2 & 1.2 \\
    \midrule
    800 & OCR & RoBERTa & 74.9 $\pm$ 4.9 & 0.0 $\pm$ 0.0 & 66.8 &  0.0 \\
    800 & OCR & Mistral-7B-Class & 78.3 $\pm$ 2.4 & 0.0 $\pm$ 0.0 & 58.7 & 0.0 \\
    800 & OCR & Mistral-7B-Gen & \textbf{79.5} $\pm$ \textbf{3.3} & 0.5 $\pm$ 0.5 & 72.8 & 0.2 \\
    800 & Image & Donut & 70.1 $\pm$ 2.6 & 0.1 $\pm$ 0.2 & 71.4 & 0.2 \\
    \midrule
    1\,600 & OCR & RoBERTa & 78.0 $\pm$ 2.0 & 0.0 $\pm$ 0.0 & 69.3 & 0.0 \\
    1\,600 & OCR & Mistral-7B-Class & \textbf{83.4} $\pm$ \textbf{4.3} & 0.0 $\pm$ 0.0 & 66.6 & 0.0 \\
    1\,600 & OCR & Mistral-7B-Gen & 82.4 $\pm$ 2.1 & 0.5 $\pm$ 0.3 & 74.7 & 0.3 \\
    1\,600 & Image & Donut & 73.8 $\pm$ 1.9 & 0.0 $\pm$ 0.0 & 76.4 & 0.1 \\
    \bottomrule
  \end{tabular}
\end{table}

One model stands out with an excellent performance for few-shot learning: Generative fine-tuning of Mistral-7B. Even when providing only $160$ training samples (10 samples per class), the LLM achieves a promising mean accuracy of $72.5\%$ on the RVL-CDIP-160x5 test sets, creating a noticeable gap to the next best model (RoBERTa, $59.8\%$). Classification fine-tuning is less successful when considering only 160 training samples but catches up for 800 and more, achieving the overall best result of $83.4\%$ for $1\,600$ training samples.

Besides the aforementioned results, \autoref{tab:finetuning} also contains the results for the original RVL-CDIP-40K test set. Regarding the OCR-based models, the results show a systematic decrease in accuracy for RVL-CDIP-40K when compared to RVL-CDIP-160x5. This is most likely due to the rather low OCR quality of the original dataset. In contrast, we are considering state-of-the-art OCR results for RVL-CDIP-160x5.

This hypothesis is further strengthened when taking the OCR-free (Donut) results into account, which exhibit a much smaller difference between the two test scenarios. Concretely, for image-based classification with Donut the RVL-CDIP-40K test results are within one or two standard deviations of the RVL-CDIP-160x5 test results for all three training scenarios (160, 800, $1\,600$). Therefore, we conclude that the five-fold selection of 160 test samples is, indeed, representative for evaluating the RVL-CDIP classification challenge.

Even though the Mistral-7B-Class achieved the single highest result on the RVL-CDIP-160x5 test sets, the generative model (Mistral-7B-Gen) is much more consistent across multiple scenarios. Given that the generative model retains its flexibility, it makes it even more appealing than using a model with a classifier head.

\subsection{Results Summary}
\label{sec:expe_summary}

A summary of the classification results is given in \autoref{tab:summary} with the best models for each training scenario (zero-shot prompting or few-shot learning) and input modality (OCR text and/or image). In the case of OCR text, we include both the embedding approach and end-to-end models. The results achieved on the RVL-CDIP-160x5 test sets are put into context with the results reported in the literature on RVL-CDIP-40K that use the full 320\,000 documents for the training.

\begin{table}[ht]
    \setlength{\tabcolsep}{10pt}
    \def\arraystretch{1.3}
    \centering
    \caption{\textbf{Summary of document classification results}. The best-performing approach is listed for each zero-shot to few-shot training scenario evaluated on the RVL-CDIP-160x5 test sets. The results are put into context with fully trained models evaluated on the RVL-CDIP-40K test set.}\label{tab:summary}
    \vspace{0.5em}
    \begin{tabular}{r l l r}
\toprule
\#Train & Input & Model & Accuracy\\
\midrule
0 & OCR & GPT-4 & 61.8 $\pm$ 2.0 \\
0 & Image & GPT-4-Vision & \textbf{69.9} $\pm$ \textbf{2.0} \\
0 & OCR+Image & GPT-4-Vision & 69.4 $\pm$ 1.7 \\
\midrule
160 & OCR & Mistral-embed+KNN & 56.4 $\pm$ 3.0 \\
160 & OCR & Mistral-7B-Gen & \textbf{72.5} $\pm$ \textbf{3.9} \\
160 & Image & Donut & 42.8 $\pm$ 3.0 \\
\midrule
800 & OCR & OpenAI-large+KNN & 64.8 $\pm$ 3.1\\
800 & OCR & Mistral-7B-Gen & \textbf{79.5} $\pm$ \textbf{3.3} \\
800 & Image & Donut & 70.1 $\pm$ 2.6 \\
\midrule
1\,600 & OCR & OpenAI-large+KNN & 67.8 $\pm$ 3.8 \\
1\,600 & OCR & Mistral-7B-Class & \textbf{83.4} $\pm$ \textbf{4.3} \\
1\,600 & Image & Donut & 73.8 $\pm$ 1.9  \\
\midrule
320\,000 & OCR & BERT~\cite{bakkali2020visual} & 85.0 \\
320\,000 & Image & Donut~\cite{kim2022donut} & 95.3 \\
320\,000 & OCR+Image & BERT+NasNet~\cite{bakkali2020visual} & 97.1 \\
\bottomrule
    \end{tabular}
\end{table}

For zero-shot prompting, the largest LLMs from OpenAI, in particular the multi-modal GPT-4-Vision model, demonstrate an impressive generalization capability with a mean accuracy of $69.9\%$ on the test sets, considering the fact that in this scenario the document classes can be changed on the fly, without the need to annotate learning samples.

Regarding the fine-tuning, the smaller Mistral-7B model stands out in its capability to rapidly adapt to the classification task with only very few training samples when using the generative fine-tuning based on LoRA. Fine-tuning with ten samples per class leads to a mean accuracy of $72.5\%$.

When fine-tuning with 100 samples per class, which is still considered to be a small amount of training data for the document classification task, the fine-tuned Mistral-7B model with a classifier head achieves the overall best mean accuracy of $83.4\%$. This is a notable achievement when compared to the $85.0\%$ accuracy reported in~\cite{bakkali2020visual} for a fully trained BERT model using $320\,000$ training samples. Admittedly, Mistral-7B is an order of magnitude larger than BERT in terms of parameters, nevertheless, it indicates that very promising results can be achieved with much less training data.

\section{Conclusion}
\label{sec:conclusion}

The RVL-CDIP dataset was originally proposed as a document classification challenge using hundreds of thousands of training samples. Under these conditions, the state of the art has gradually achieved near-perfect performance. By revisiting the question of document classification under the perspective of considering only very few training samples (or none at all), this paper investigates the capacity of current document models to rapidly generalize to new tasks.

We contribute a comprehensive set of benchmark results that explore the question with prompts, embeddings, and model fine-tuning using methods from the current state of the art for image and text analysis. We demonstrate the feasibility of zero-shot and few-shot document classification using LLMs, achieving results that, although promising between $69.9\%$ and $83.4\%$ mean accuracy, leave significant room for improvement.

An important line of future research is related to document foundation models. The strongest few-shot fine-tuning results reported in this paper were achieved with large text-based models (Mistral-7B). However, the state of the art clearly demonstrates that combining image and text leads to significantly better results for fully trained models. Therefore, integrating more visual information into document foundation models is expected to significantly improve few-shot document classification. There is an increasing number of multi-modal LLMs~\cite{mmllms} that may be investigated in this context.

Other lines of research include improvements of the prompts, e.g. by providing additional semantics to the LLM that go beyond only the name of the category. Finally, it would be beneficial to explore different learning strategies for unlabeled data, including self-training and unsupervised contrastive learning, to further improve the results of few-shot fine-tuning.

\bibliographystyle{acm}
\bibliography{biblio}

\begin{thebibliography}{10}

\bibitem{achiam2023gpt}
{\sc Achiam, J., Adler, S., Agarwal, S., Ahmad, L., Akkaya, I., Aleman, F.~L.,
  Almeida, D., Altenschmidt, J., Altman, S., Anadkat, S., et~al.}
\newblock Gpt-4 technical report.
\newblock {\em arXiv preprint arXiv:2303.08774\/} (2023).

\bibitem{bakkali2020visual}
{\sc Bakkali, S., Ming, Z., Coustaty, M., and Rusi{\~n}ol, M.}
\newblock Visual and textual deep feature fusion for document image
  classification.
\newblock In {\em Proc. Int. Conf. on Computer Vision and Pattern Recognition
  Workshops (CVPRW)\/} (2020), pp.~562--563.

\bibitem{brown2020language}
{\sc Brown, T., Mann, B., Ryder, N., Subbiah, M., Kaplan, J.~D., Dhariwal, P.,
  Neelakantan, A., Shyam, P., Sastry, G., Askell, A., et~al.}
\newblock Language models are few-shot learners.
\newblock {\em Advances in neural information processing systems (NeurIPS)
  33\/} (2020), 1877--1901.

\bibitem{qlora}
{\sc Dettmers, T., Pagnoni, A., Holtzman, A., and Zettlemoyer, L.}
\newblock Qlora: Efficient finetuning of quantized llms.
\newblock {\em Advances in Neural Information Processing Systems 36\/} (2024).

\bibitem{gunther2023jina}
{\sc G{\"u}nther, M., Ong, J., Mohr, I., Abdessalem, A., Abel, T., Akram,
  M.~K., Guzman, S., Mastrapas, G., Sturua, S., Wang, B., et~al.}
\newblock Jina embeddings 2: 8192-token general-purpose text embeddings for
  long documents.
\newblock {\em arXiv preprint arXiv:2310.19923\/} (2023).

\bibitem{harley2015icdar}
{\sc Harley, A.~W., Ufkes, A., and Derpanis, K.~G.}
\newblock Evaluation of deep convolutional nets for document image
  classification and retrieval.
\newblock In {\em Proc. Int. Conf. on Document Analysis and Recognition
  (ICDAR)\/} (2015), pp.~991--995.

\bibitem{lora}
{\sc Hu, E.~J., Shen, Y., Wallis, P., Allen-Zhu, Z., Li, Y., Wang, S., Wang,
  L., and Chen, W.}
\newblock Lora: Low-rank adaptation of large language models.
\newblock {\em arXiv preprint arXiv:2106.09685\/} (2021).

\bibitem{jiang2023mistral}
{\sc Jiang, A.~Q., Sablayrolles, A., Mensch, A., Bamford, C., Chaplot, D.~S.,
  Casas, D. d.~l., Bressand, F., Lengyel, G., Lample, G., Saulnier, L., et~al.}
\newblock Mistral 7b.
\newblock {\em arXiv preprint arXiv:2310.06825\/} (2023).

\bibitem{mistral}
{\sc Jiang, A.~Q., Sablayrolles, A., Mensch, A., Bamford, C., Chaplot, D.~S.,
  Casas, D. d.~l., Bressand, F., Lengyel, G., Lample, G., Saulnier, L., et~al.}
\newblock Mistral 7b.
\newblock {\em arXiv preprint arXiv:2310.06825\/} (2023).

\bibitem{jiang2024mixtral}
{\sc Jiang, A.~Q., Sablayrolles, A., Roux, A., Mensch, A., Savary, B., Bamford,
  C., Chaplot, D.~S., de~las Casas, D., Hanna, E.~B., Bressand, F., Lengyel,
  G., Bour, G., Lample, G., Lavaud, L.~R., Saulnier, L., Lachaux, M.-A., Stock,
  P., Subramanian, S., Yang, S., Antoniak, S., Scao, T.~L., Gervet, T., Lavril,
  T., Wang, T., Lacroix, T., and Sayed, W.~E.}
\newblock Mixtral of experts.
\newblock {\em arXiv preprint arXiv:2401.04088\/} (2024).

\bibitem{kim2022donut}
{\sc Kim, G., Hong, T., Yim, M., Nam, J., Park, J., Yim, J., Hwang, W., Yun,
  S., Han, D., and Park, S.}
\newblock Ocr-free document understanding transformer.
\newblock In {\em Proc. European Conference on Computer Vision (ECCV)\/}
  (2022), Springer, pp.~498--517.

\bibitem{lewis2006building}
{\sc Lewis, D., Agam, G., Argamon, S., Frieder, O., Grossman, D., and Heard,
  J.}
\newblock Building a test collection for complex document information
  processing.
\newblock In {\em Proceedings of the 29th annual international ACM SIGIR
  conference on Research and development in information retrieval\/} (2006),
  pp.~665--666.

\bibitem{roberta}
{\sc Liu, Y., Ott, M., Goyal, N., Du, J., Joshi, M., Chen, D., Levy, O., Lewis,
  M., Zettlemoyer, L., and Stoyanov, V.}
\newblock Roberta: A robustly optimized bert pretraining approach.
\newblock {\em arXiv preprint arXiv:1907.11692\/} (2019).

\bibitem{neelakantan2022text}
{\sc Neelakantan, A., Xu, T., Puri, R., Radford, A., Han, J.~M., Tworek, J.,
  Yuan, Q., Tezak, N., Kim, J.~W., Hallacy, C., et~al.}
\newblock Text and code embeddings by contrastive pre-training.
\newblock {\em arXiv preprint arXiv:2201.10005\/} (2022).

\bibitem{park2019cord}
{\sc Park, S., Shin, S., Lee, B., Lee, J., Surh, J., Seo, M., and Lee, H.}
\newblock Cord: A consolidated receipt dataset for post-ocr parsing.
\newblock In {\em Document Intelligence Workshop at Neural Information
  Processing Systems\/} (2019).

\bibitem{radford2018improving}
{\sc Radford, A., Narasimhan, K., Salimans, T., Sutskever, I., et~al.}
\newblock Improving language understanding by generative pre-training.
\newblock {\em OpenAI\/} (2018).

\bibitem{radford2019language}
{\sc Radford, A., Wu, J., Child, R., Luan, D., Amodei, D., Sutskever, I.,
  et~al.}
\newblock Language models are unsupervised multitask learners.
\newblock {\em OpenAI blog 1}, 8 (2019), 9.

\bibitem{raffel2020exploring}
{\sc Raffel, C., Shazeer, N., Roberts, A., Lee, K., Narang, S., Matena, M.,
  Zhou, Y., Li, W., and Liu, P.~J.}
\newblock Exploring the limits of transfer learning with a unified text-to-text
  transformer.
\newblock {\em Journal of machine learning research 21}, 140 (2020), 1--67.

\bibitem{tensmeyer2017analysis}
{\sc Tensmeyer, C., and Martinez, T.}
\newblock Analysis of convolutional neural networks for document image
  classification.
\newblock In {\em Proc. Int. Conf. on Document Analysis and Recognition
  (ICDAR)\/} (2017), pp.~388--393.

\bibitem{hf-transformers}
{\sc Wolf, T., Debut, L., Sanh, V., Chaumond, J., Delangue, C., Moi, A.,
  Cistac, P., Rault, T., Louf, R., Funtowicz, M., et~al.}
\newblock Huggingface's transformers: State-of-the-art natural language
  processing.
\newblock {\em arXiv preprint arXiv:1910.03771\/} (2019).

\bibitem{xu2021layoutlmv2}
{\sc Xu, Y., Xu, Y., Lv, T., Cui, L., Wei, F., Wang, G., Lu, Y., Florencio, D.,
  Zhang, C., Che, W., Zhang, M., and Zhou, L.}
\newblock {L}ayout{LM}v2: Multi-modal pre-training for visually-rich document
  understanding.
\newblock In {\em Proc. Int. Joint Conference on Natural Language Processing
  (IJCNLP)\/} (2021), pp.~2579--2591.

\bibitem{mmllms}
{\sc Zhang, D., Yu, Y., Dong, J., Li, C., Su, D., Chu, C., and Yu, D.}
\newblock {MM-LLMs:} recent advances in multimodal large language models.
\newblock {\em arXiv preprint arXiv:2401.13601\/} (2024).

\end{thebibliography}

\end{document}